\documentclass{article}
\usepackage{ijcai25}
\usepackage{placeins}
\usepackage{times}
\usepackage{soul}
\usepackage{url}
\usepackage[hidelinks]{hyperref}
\usepackage[utf8]{inputenc}
\usepackage[small]{caption}
\usepackage{graphicx}
\usepackage{amsmath}
\usepackage{amsthm}
\usepackage{booktabs}
\usepackage{algorithm}
\usepackage{algorithmic}
\usepackage[switch]{lineno}

\usepackage{tabularx}
\usepackage{makecell}
\usepackage{listings}
\usepackage{tcolorbox}
\usepackage{amsmath}
\usepackage{xcolor}

\usepackage{multirow}

\usepackage[belowskip=2pt,aboveskip=0pt]{caption}

\title{Resolving Conflicting Evidence in Automated Fact-Checking: \\A Study on Retrieval-Augmented LLMs}

\author{
Ziyu Ge$^{1}$\footnote{These authors contributed equally.}\and
Yuhao Wu$^{1*}$\and
Daniel Wai Kit Chin$^1$\and
Roy Ka-Wei Lee$^1$\And
Rui Cao$^2$\\
\affiliations
$^1$Singapore University of Technology and Design\\
$^2$University of Cambridge\\
\emails
\{ziyu\_ge, roy\_lee\}@sutd.edu.sg,
\{yuhao\_wu, daniel\_chin\}@mymail.sutd.edu.sg,
rc990@cam.ac.uk
}

\begin{document}

\maketitle

\begin{abstract}
    Large Language Models (LLMs) augmented with retrieval mechanisms have demonstrated significant potential in fact-checking tasks by integrating external knowledge. However, their reliability decreases when confronted with conflicting evidence from sources of varying credibility. This paper presents the first systematic evaluation of Retrieval-Augmented Generation (RAG) models for fact-checking in the presence of conflicting evidence. To support this study, we introduce \textbf{CONFACT} (\textbf{Con}flicting Evidence for \textbf{Fact}-Checking)\footnote{Dataset available at \url{https://github.com/zoeyyes/CONFACT}}, a novel dataset comprising questions paired with conflicting information from various sources. Extensive experiments reveal critical vulnerabilities in state-of-the-art RAG methods, particularly in resolving conflicts stemming from differences in media source credibility. To address these challenges, we investigate strategies to integrate media background information into both the retrieval and generation stages. Our results show that effectively incorporating source credibility significantly enhances the ability of RAG models to resolve conflicting evidence and improve fact-checking performance.
\end{abstract}

\section{Introduction}
\paragraph{Motivation.} Fact-checking systems are essential tools for combating the spread of misinformation, as they help verify claims by retrieving and analyzing evidence from diverse sources~\cite{DBLP:journals/tacl/GuoSV22,DBLP:conf/ijcai/NakovCHAEBPSM21}. Modern fact-checking pipelines increasingly rely on Retrieval-Augmented Generation (RAG) frameworks, which integrate external evidence into Large Language Models (LLMs) to verify claims~\cite{DBLP:conf/nips/LewisPPPKGKLYR020,guu2020retrieval}. However, a critical challenge arises when fact-checking systems encounter \textit{conflicting evidence}—that is, when retrieved documents present opposing stances on a claim, often originating from sources with varying levels of credibility~\cite{DBLP:journals/tacl/GuoSV22,DBLP:journals/corr/abs-2409-00781,DBLP:conf/naacl/HongKKMW24}.

\begin{figure}[t] 
	\centering
    	
	\includegraphics[width=0.9\linewidth]{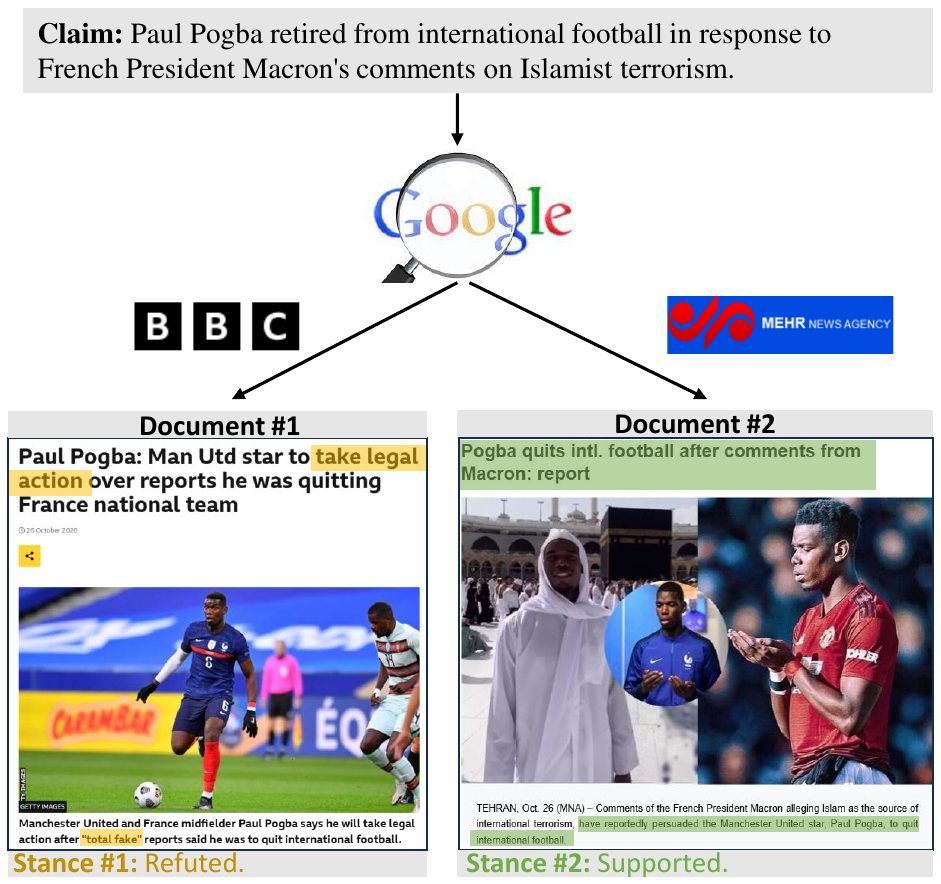} 
	\caption{
    The retrieved documents from Google to verify the claim.
    The retrieved documents from different media sources have different stances towards the claim.
    }
	\label{fig:intro-img}
\end{figure}

For example, consider the claim: ``\textit{Paul Pogba retired from international football in response to French President Macron's comments on Islamist terrorism}'', the retrieved evidence might include conflicting documents, as shown in Figure~\ref{fig:intro-img}, such as one from BBC\footnote{\url{https://www.bbc.co.uk/sport/football/54691842}}, a highly credible source, and another from Mehr News Agency\footnote{\url{https://en.mehrnews.com/news/165168/Pogba-quits-intl-football-after-comments-from-Macron-report}}, which is flagged as untrustworthy\footnote{\url{https://mediabiasfactcheck.com/mehr-news-agency/}}. To fact-check this claim accurately, a fact-checking system must not only analyze the evidence but also assess the credibility of each source—prioritizing reliable information while discounting less trustworthy content.

This challenge is exacerbated by the rapid proliferation of low-credibility content and automated misinformation generated by LLMs themselves~\cite{DBLP:conf/iclr/ChenS24,DBLP:journals/corr/abs-2408-11871}. Fact-checking in this context requires robust systems capable of resolving conflicts in evidence while reasoning about source credibility—capabilities that are currently underexplored in fact-checking research.

\paragraph{Research Objectives.} Addressing these gaps, this paper focuses on the problem of \textit{fact-checking with conflicting evidence}, where retrieved documents present opposing stances on a claim. Specifically, we aim to evaluate the ability of retrieval-augmented LLMs to identify, analyze, and resolve conflicts in evidence by determining which sources to trust for claim verification. To enable this, we introduce \textsf{CONFACT} (\textbf{Con}flicting Evidence for \textbf{Fact}-Checking), a novel dataset designed to systematically study this challenge.  Each instance in CONFACT comprises a claim paired with documents exhibiting conflicting stances, annotated with source credibility ratings.

We conduct extensive experiments to evaluate state-of-the-art RAG models on CONFACT, revealing critical limitations in their ability to reason through conflicting evidence and prioritize trustworthy sources. Motivated by these findings, we further explore strategies for incorporating media background information—such as source metadata and credibility scores—into both the retrieval and generation processes. Our results demonstrate that effectively integrating source credibility enhances the robustness of retrieval-augmented LLMs in resolving conflicting evidence for fact-checking.

\paragraph{Contributions.}  
In this work, we made the following contributions in this work:
\begin{itemize}
    \item \textbf{Dataset Creation}: We introduce \textsf{CONFACT}, a novel dataset for studying fact-checking with conflicting evidence. The dataset includes claims paired with conflicting retrieved documents, annotated with source credibility and stance labels to facilitate systematic evaluation.
    \item \textbf{Performance Evaluation}: We conduct a comprehensive evaluation of RAG-based LLMs on \textsf{CONFACT}, revealing critical vulnerabilities in resolving conflicting evidence and reasoning about source credibility.
    \item \textbf{Methodological Innovations}: We propose and evaluate multiple strategies for integrating media background information into RAG pipelines, demonstrating significant improvements in fact-checking performance through effective credibility-aware reasoning.
\end{itemize}

\section{Related Work}
\subsection{RAG for Automated Fact-Checking}
\label{sec:related-rag}
Automated fact-checking (AFC) has gained significant attention in recent years~\cite{DBLP:journals/tacl/GuoSV22,DBLP:conf/ijcai/NakovCHAEBPSM21}.
While LLMs have demonstrated strong performance in various Natural Language Understanding (NLU) tasks~\cite{DBLP:journals/corr/abs-2311-03731}, they remain limited in AFC, as fact-checking often requires evidence beyond the parametric knowledge stored within LLMs~\cite{DBLP:conf/nips/SchlichtkrullG023,thorne-etal-2018-fever,wang-2017-liar}.
RAG~\cite{DBLP:conf/nips/LewisPPPKGKLYR020,DBLP:journals/tacl/RamLDMSLS23} facilitates the adaptation of LLMs to AFC by incorporating external retrieved evidence to LLMs~\cite{DBLP:conf/emnlp/PanLKN23,DBLP:conf/acl/PanWLLWKN23,DBLP:conf/naacl/ChenKSDC24,zhang-gao-2024-reinforcement}.
However, not all retrieved evidence is reliable~\cite{DBLP:journals/tacl/GuoSV22,DBLP:conf/naacl/HongKKMW24}, and information from untrustworthy sources may contain misinformation, leading to conflicting evidence.
Recent studies have shown that retrieval-augmented LLMs are particularly vulnerable to contradictions in augmented texts~\cite{DBLP:conf/emnlp/MinMHZ20,DBLP:journals/corr/abs-2404-12447,DBLP:conf/acl/ChenGLL020,DBLP:conf/acl/AmplayoW00N23}. 
Given the risks posed by unreliable sources, it is crucial to investigate the robustness of retrieval-augmented LLMs in AFC, particularly in handling conflicting evidence.

\subsection{Source Credibility Estimation}
\label{sec:related-source-credb}
Source credibility estimation is crucial, as not all media sources are reliable; however, this problem remains underexplored. Early works addressed this issue by estimating media credibility through analysis of fake news records associated with sources~\cite{DBLP:conf/cikm/MukherjeeW15,DBLP:conf/cikm/PopatMSW16,DBLP:conf/www/PopatMSW17}. The authors in~\cite{DBLP:conf/emnlp/BalyKAGN18} introduced the first dataset with human-annotated factuality ratings of news sources and utilized various features, such as Wikipedia information and source URLs, for credibility estimation. Subsequent studies proposed more robust models using diverse features of media sources~\cite{DBLP:conf/emnlp/ZhangMBRAKSJKBD19,DBLP:conf/acl/BalyKAKDAGN20,DBLP:conf/uss/HounselHKBFM20}. In contrast to these classification approaches, the work in~\cite{DBLP:journals/corr/abs-2409-00781} emphasized the generation of detailed background checks for media sources.

Despite these advancements, the impact of estimated source credibility in fact-checking is still unknown.
has received limited attention. 
To date, only~\cite{DBLP:journals/corr/abs-2409-00781} conducted a small-scale experiment with 20 claims, examining whether incorporating source background checks could
benefit claim verification.
In this paper, we extend this line of inquiry by comprehensively evaluating how media source backgrounds can facilitate fact-checking models, and exploring optimal strategies for integrating source credibility information into these systems.

\section{CONFACT Dataset}
The \textsf{CONFACT} dataset is specifically designed to facilitate the study of fact-checking in scenarios where conflicting evidence is retrieved from sources of varying credibility, thereby addressing a critical gap in existing datasets like AVERITEC~\cite{DBLP:conf/nips/SchlichtkrullG023} and FactCheckQA~\cite{DBLP:journals/corr/abs-2311-06697}. The construction involved two key steps: 1) identifying claims likely to retrieve conflicting evidence -- particularly those frequently associated with misinformation from untrustworthy sources, and 2) ensuring that retrieved documents for claim verification present conflicting stances.

\subsection{Claim Collection}
\label{sec:dataset-data}
To identify claims likely to retrieve conflicting evidence, we utilized two widely used fact-checking datasets:
\begin{itemize}
    \item \textbf{AVERITEC.} This dataset~\cite{DBLP:conf/nips/SchlichtkrullG023} contains 4,568 real-world claims fact-checked by 50 organizations, categorized as \textit{Conflicting Evidence/Cherry-picking}, \textit{Not Enough Evidence}, \textit{Refuted}, and \textit{Supported}. We selected claims labeled as \textit{Refuted} or \textit{Supported}, which involve clear factuality.
    \item \textbf{FactCheckQA.} This dataset~\cite{DBLP:journals/corr/abs-2311-06697} includes 20,871 claims annotated as \textit{true}, \textit{false}, or \textit{other}. We focused on claims labeled as \textit{true} or \textit{false}, which provide definitive factuality.
\end{itemize}

Claims from these datasets were merged\footnote{Claims from social media platforms were excluded 
as they are less findable by search engines.}, covering diverse topics. 
This process resulted in 3,180 claims: 566 from AVERITEC and 2,614 from FactCheckQA.

\subsection{Conflicting Evidence Collection}
\label{sec:dataset-conflicts}
To facilitate the study of conflicting evidence in fact-checking, we retrieved relevant documents for claim verification. Instead of directly querying Google with the original claims, we transformed each claim into a binary question regarding its veracity using GPT-4\footnote{https://openai.com/index/gpt-4/}, following the approach outlined in~\cite{DBLP:journals/corr/abs-2409-00781,DBLP:journals/corr/abs-2311-06697}. For example, the claim \textit{Nigeria had a population of 45 million at the time of independence''} was converted into the question \textit{Did Nigeria have a population of 45 million at the time of independence?''}. Each question was then submitted as a query on Google, from which we retrieved the top 10 web pages\footnote{Searches and scraping were conducted within a single week (September 12--19, 2024)}. To ensure reproducibility, the retrieved web pages were archived using the Wayback Machine\footnote{https://web.archive.org/}.

\subsection{Conflict Evidence Annotation}
Next, we annotated the stances of the collected evidence documents using a two-stage process designed to identify conflicting viewpoints.

\textbf{Stage 1: GPT-4 Annotation.}  
We employed GPT-4 to classify the stance of each document with respect to its corresponding claim as either \textit{supporting} or \textit{refuting}. To enhance robustness, we used three distinct prompt variations: (i) classify the stance based solely on the document URL; (ii) classify the stance using the retrieved webpage content; and (iii) prompt GPT-4 to provide its reasoning prior to making a classification. 
The specific prompts are detailed in Appendix~\ref{sec:app-prompt-annotation}. 

The final stance for each document was determined through majority voting across these three approaches. We defined a claim as exhibiting conflicting evidence if it was associated with documents classified as both \textit{supporting} and \textit{refuting}. Out of 3,180 claims, 611 (17.8\%) met this criterion and advanced to the next stage.

\textbf{Stage 2: Human Annotation.}  
Human annotators subsequently validated the conflicting evidence identified in Stage 1. For each claim, annotators reviewed pairs of documents---one labeled as \textit{supporting} and another as \textit{refuting} by GPT-4. The annotators verified the stances and assessed the credibility of the sources on a 5-point scale (1 = least credible, 5 = most credible). Additionally, they categorized each source into one of the following groups: \textit{Mainstream News}, \textit{Government}, \textit{Non-profit}, \textit{Academic}, \textit{Social Media}, or \textit{Other}. Each document pair was independently reviewed by two annotators, with any disagreements resolved by a third annotator. Detailed annotation guidelines are provided in Appendix~\ref{sec:app-annotation-guide}.

\begin{table}[t]
\centering
  \begin{tabular}{lcc}
    \toprule
    \textbf{Split}  & \textbf{Labels}& \textbf{\# Sources} \\ \midrule
    ModC &125 Yes; 486 No & 2469\\ 
    HumC &51 Yes; 236 No & 1418 \\
    \bottomrule
\end{tabular}
\caption{Statistics of the ModC and HumC split of our \textsf{CONFACT}.}
  \label{tab:split-dist}
\end{table}
\subsection{Dataset Analysis}
\label{sec:dataset-analysis}
The final \textsf{CONFACT} dataset consists of two splits: \textit{Model Conflicts} (ModC) and \textit{Human Conflicts} (HumC).
ModC comprises claims with conflicting evidence identified by GPT-4 during Stage 1. Given that GPT-4 is a powerful closed-source model, this split contains conflicts that may be particularly challenging for most open-source models to resolve.
HumC consists of claims where the evidence is conflicting from a human perspective, aiming to assess how effectively fact-checking systems can mitigate human uncertainty when verifying such evidence. The inter-annotator agreement for HumC, as measured by Krippendorff’s Alpha, was 0.586—indicating strong agreement while also reflecting the general confusion among annotators when dealing with conflicting documents.
Following prior work~\cite{DBLP:journals/corr/abs-2311-06697,DBLP:journals/corr/abs-2409-00781}, we further formulate the claim verification task into a binary question regarding claim veracity, making it more naturally suited for retrieval-augmented LLMs. The binary questions were generated with GPT-4 as discussed in Section~\ref{sec:dataset-conflicts}.
Claims labeled as \textit{true/supported} correspond to questions with \textit{Yes} as answers, and those labeled as \textit{false/refuted} correspond to questions with
\textit{No} as answers. The statistics of  \textsf{CONFACT} are provided in Table~\ref{tab:split-dist}  and an illustration of a data sample from \textsf{CONFACT} is provided in Appendix~\ref{sec:app-illu-data}.

An analysis of document credibility revealed key challenges in assessing source credibility. Annotators frequently overestimated the reliability of \textit{Mainstream News} sources, with 95.8\% of these sources rated as credible or neutral. Cross-referencing these ratings with expert annotations from Media Bias / Fact Check (MBFC)\footnote{https://mediabiasfactcheck.com/} showed that 30\% of misleading sources were flagged as unreliable by MBFC, while annotators classified 69.8\% of these as trustworthy. These findings underscore the challenges of accurately assessing credibility and highlight the importance of addressing conflicting evidence in fact-checking tasks. 
More details for the distribution of source credibility over source types are available in Appendix~\ref{sec:app-dis-source-cred}.

\section{Methodology}
In this section, we evaluate retrieval-augmented LLMs on the \textsf{CONFACT} dataset to assess their robustness in fact-checking when confronted with conflicting evidence. We begin by formally defining the task in Section~\ref{sec:method-prob-def}. Next, in Section~\ref{sec:method-vanilla-rag}, we describe baseline retrieval-augmented LLMs for fact-checking. Finally, Section~\ref{sec:method-rag-backgrounds} presents strategies for incorporating media source background information at various stages of the RAG pipeline.

\subsection{Problem Definition}
\label{sec:method-prob-def}
Given a claim verification question $\mathcal{Q}$ with its relevant $N$ retrieved documents $\{\mathcal{D}_n\}_{n=1}^N$ from \textsf{CONFACT}, 
a retrieval-augmented LLM is expected to 
generate an answer $\mathcal{A}$ to the question that reflects the veracity of the original claim against available evidence. 
The system is evaluated on its accuracy in correctly predicting the veracity of claims (i.e., whether $\mathcal{A}$ exactly matches the ground-truth label $\hat{\mathcal{A}}$ for the converted question for claim verification). In addition, we report the \textit{Macro-F1} score as an auxiliary metric to assess performance across classes, particularly given the imbalanced nature of the dataset.

\begin{figure}[t] 
	\centering
	\includegraphics[width=0.9\linewidth]{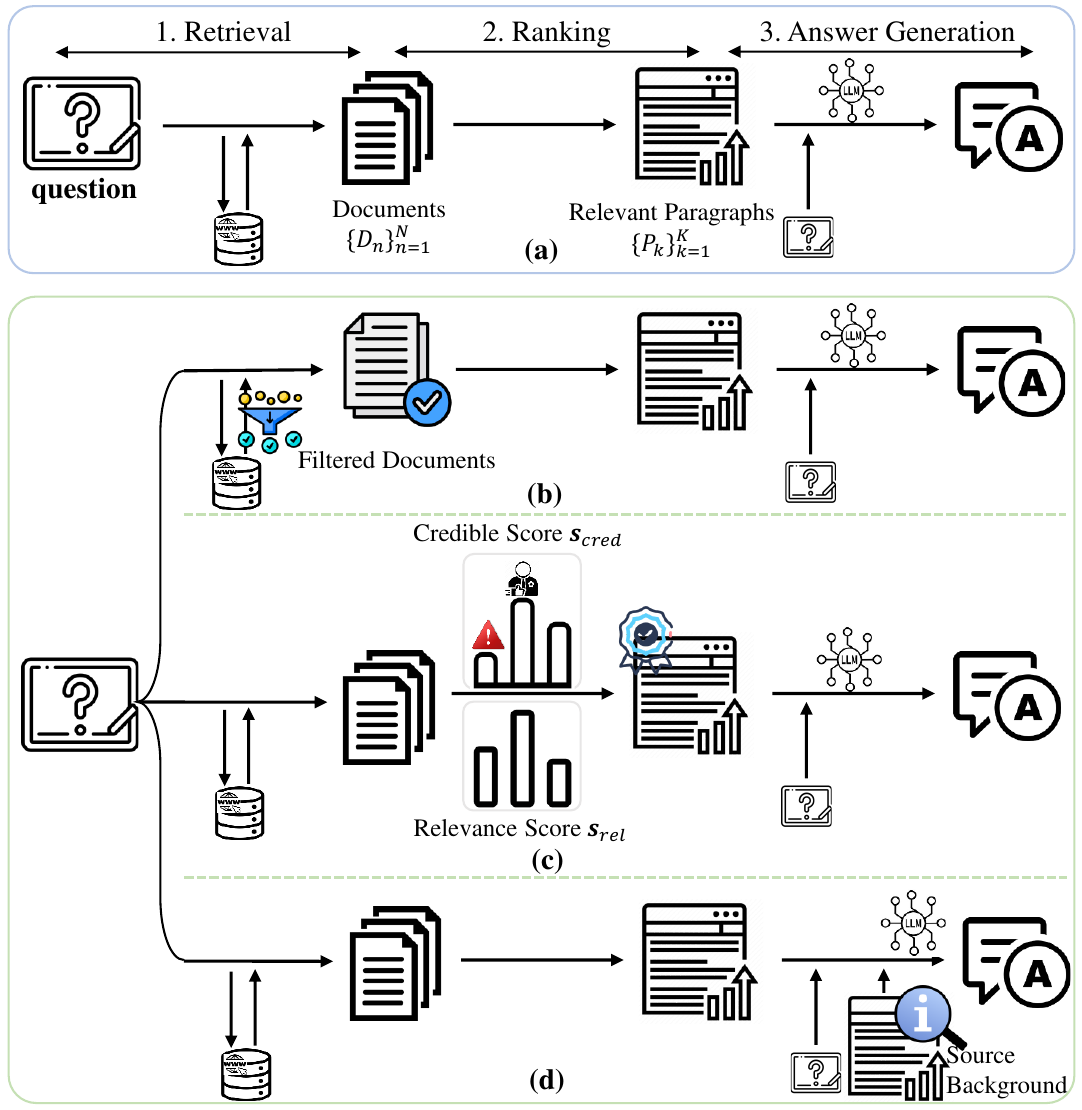} 
	\caption{(a) illustrates a general framework of RAG methods involving three stages: retrieval, ranking and answer generation. (b-d) demonstrate our source-aware retrieval-augmented LLMs, incorporating source background information in three stages of the general RAG framework.}
	\label{fig:model-rag-img}
\end{figure}

A typical retrieval-augmented fact-checking workflow consists of three main stages: \textit{retrieval}, \textit{ranking}, and \textit{answer generation}~\cite{DBLP:conf/emnlp/WangWGZWXSWLQYL24,DBLP:journals/corr/abs-2312-10997}, as illustrated in Figure~\ref{fig:model-rag-img}(a).

\begin{itemize}
    \item \textit{Retrieval}: Given a claim verification question $\mathcal{Q}$, a RAG model retrieves relevant documents from an external knowledge base, represented as $\{\mathcal{D}_n\}_{n=1}^N$. Retrieved documents were provided on \textsf{CONFACT} to ensure reproducibility, as retrieval is time-varying.
    \item \textit{Ranking}: Retrieved documents are chunked into short passages, and a ranking function selects the top-$K$ most relevant paragraphs $\{\mathcal{P}_k\}_{k=1}^K$ for fact-checking.
    \item \textit{Answer Generation}: The selected paragraphs $\{\mathcal{P}_k\}_{k=1}^K$ are passed to an LLM to generate the final answer $\mathcal{A}$.
\end{itemize}

\subsection{Baseline Retrieval-Augmented LLMs}
\label{sec:method-vanilla-rag}
Baseline retrieval-augmented LLMs adhere to the standard workflow illustrated in Figure~\ref{fig:model-rag-img}(a). In this process, the most relevant set of paragraphs $\{\mathcal{P}_k\}_{k=1}^K$ are extracted and used as input for answer generation. We evaluate multiple prompting strategies for leveraging these augmented contexts:

\begin{itemize}
    \item \textit{Direct Answer} (\textbf{DirA.}):  The $K$ selected paragraphs are provided to the LLM along with the claim verification question, and the model directly generates an answer.
    \item \textit{Majority Vote} (\textbf{MajV.}): The model first predicts answer candidates $\mathcal{A}_k$ for each paragraph $\mathcal{P}_k$. A majority vote is then conducted to select the final answer.
    \item \textit{Discern and Answer} (\textbf{DisA.}): Inspired by~\cite{DBLP:conf/naacl/HongKKMW24}, an explicit instruction is added to filter out misleading passages before generating an answer.
    \item \textit{Chain-of-Thought} (\textbf{CoT}): This strategy prompts the LLM to generate a rationale before predicting the answer~\cite{DBLP:conf/nips/Wei0SBIXCLZ22}, improving reasoning in multi-step verification tasks.
\end{itemize}

While these strategies perform well in standard question-answering tasks, they struggle when the retrieved evidence exhibits conflicting viewpoints. For instance, DirA. may conflate misinformation with factual content, MajV. fails if misleading sources outnumber reliable ones, and DisA. depends on the LLM’s ability to filter unreliable information, which is not always effective. These limitations motivate the incorporation of media background knowledge.

\subsection{Retrieval-Augmented LLMs with Media Source Backgrounds}
\label{sec:method-rag-backgrounds}

To improve fact-checking performance in the presence of conflicting evidence, we propose integrating background information from the source of the media at different stages of the RAG pipeline.

\subsubsection{Media Source Background Provider}
\label{sec:method-media-gen}

For each retrieved document, we extract background information about its source. The MBFC website
serves as our primary source background provider (GT-MB), offering expert annotations on media bias and factual reliability. If a source is available in MBFC, its credibility rating is retrieved. Otherwise, the background is marked as missing.

To extend coverage beyond MBFC, we introduce a \textbf{Hybrid-MB} provider, combining MBFC annotations with an LLM-based background generator~\cite{DBLP:journals/corr/abs-2409-00781}. The generator first retrieves real-time information about the source’s publisher, past credibility ratings, and history of misinformation via Google Search APIs\footnote{https://developers.google.com/custom-search/v1/overview}. It then processes this information using an in-context learning approach with a set of pre-defined prompts, generating a credibility summary (denoted as $\mathcal{B}$)  that
includes factual accuracy, bias, and misinformation history (Refer to Appendix~\ref{sec:app-media-background-pred} for the designed prompts). 

Although the generated source credibility description is comprehensive, it may not be directly applicable at all stages of retrieval-augmented LLMs.
Therefore, we
further map this description into a credibility score $s_\text{cred} \in (0,1)$ using a prediction model $\pi_{\theta}$:
\begin{equation}
\label{eq:credible-score}
    s_\text{cred} =\pi_{\theta}(\mathcal{B}).
\end{equation}
The model is trained on~\cite{DBLP:conf/emnlp/BalyKAGN18}, which provides labeled credibility supervision. More details about the credible score prediction are provided in Appendix~\ref{sec:app-media-background-pred}.

\subsubsection{Media Background Incorporation}
\label{sec:method-media-incorp}

We explore to incorporate source credibility information in three stages of the RAG pipeline:

\textit{1. Source Filtering in Retrieval (\textbf{SF})}:  It aims to filter incredible information in the document level. Documents from sources 
described as \textit{low credible} according to $\mathcal{B}$ 
are filtered before ranking (Figure~\ref{fig:model-rag-img}(b)) 
(more details in Appendix~\ref{sec:app-cred-ext-from-background}).
The remaining documents are ranked, and the top-$K$ paragraphs are used for answer generation.

\textit{2. Credibility Weighting in Ranking (\textbf{CW})}:  
Instead of filtering in the document level, credibility scores influence ranking (Figure~\ref{fig:model-rag-img}(c)). The final ranking score for a paragraph $\mathcal{P}_m$ is computed as:
\begin{equation}
    s_m  = s_{\text{rel},m} + \beta * s_{\text{cred},m},
\end{equation}
where $s_{\text{rel},m}$ is the relevance score and $\beta$ balances relevance and credibility.
We considered both a soft (CW$_\text{soft}$) and a hard (CW$_\text{hard}$) setting for leveraging the credible score where CW$_\text{hard}$ further maps $s_\text{cred}$ into 0 and 1. Specifically, if $s_\text{cred}$ is below a threshold $\gamma$, it will be mapped to 0, otherwise, 1.

\textit{3. Source Backgrounds Augmentation in Generation (\textbf{SBA})}:  
Source backgrounds are included at the answer generation stage (Figure~\ref{fig:model-rag-img}(d)). We evaluate four strategies:
\begin{itemize}
    \item \textbf{SBA$_\text{dir}$}: Concatenates each paragraph with its source background for source-aware paragraphs ($[\mathcal{P}_k, \mathcal{B}_k]$). The $K$ source-aware paragraphs are fed to LLMs for a direct answer. 
    \item \textbf{SBA$_\text{CoT}$}: Uses CoT prompt with source-aware paragraphs.
    \item \textbf{SBA$_\text{exp}$}: Receives  source-aware paragraphs and uses explicit instructions to filter unreliable sources.
    \item \textbf{SBA$_\text{ens}$}: Uses a two-stage process where candidate answers are generated per paragraph, and conflicts are resolved based on source-aware rationales:
\begin{equation}
\mathcal{A}_k, \mathcal{R}_k = \text{LLM}([\mathcal{P}_k, \mathcal{B}_k], \mathcal{Q})
\end{equation}
\begin{equation}
\mathcal{A}^* = \text{LLM}([\mathcal{A}_1, \mathcal{R}_1, \dots, \mathcal{A}_D, \mathcal{R}_D], \mathcal{Q})
\end{equation}
where $\mathcal{A}^*$ is the final answer after considering all rationales. 

Refer to Appendix~\ref{sec:app-src-back-incorp} for the designed prompts. 
\end{itemize}

\section{Experiments}
\begin{table*}[t]
  \centering
\small
  \begin{tabularx}{\linewidth}{XX|XXXXXX|XXXXXX}
    \hline
     & &
    \multicolumn{6}{c}{\textbf{ModC}} & \multicolumn{6}{|c}{\textbf{HumC}}\\ 
   \textbf{Set.}&\textbf{Meth.}& \multicolumn{2}{c}{\textbf{LLaMA-3.1}} & \multicolumn{2}{c}{\textbf{Qwen-2}} & \multicolumn{2}{c}{\textbf{Mistral}}  & \multicolumn{2}{|c}{\textbf{LLaMA-3.1}} & \multicolumn{2}{c}{\textbf{Qwen-2}} & \multicolumn{2}{c}{\textbf{Mistral}}    \\
 \cmidrule(lr){3-4} \cmidrule(lr){5-6}   \cmidrule(lr){7-8}   \cmidrule(lr){9-10}   \cmidrule(lr){11-12}   \cmidrule(lr){13-14}  
   && \textbf{Acc.}&   \textbf{F1}& \textbf{Acc.} &  \textbf{F1}& \textbf{Acc.} & \textbf{F1} & \textbf{Acc.}&   \textbf{F1}& \textbf{Acc.} &  \textbf{F1}& \textbf{Acc.} & \textbf{F1} \\
   \hline \hline
    \multirow{4}{*}{\textbf{Bsl.}} 
     &DirA. & 71.36 &66.13 &78.40  & 45.96 &  77.58 &  70.82  & 70.03  &63.81      &77.70  &67.83   &75.96   & 67.87  \\
     &MajV.  &79.87&45.96 &79.71& 45.90 & 79.54  &  45.83  &  82.93 & 49.07     & 82.93 & 49.07  & 82.93  &49.07    \\
     &DisA.  & 72.18 &63.26 &  78.89& 69.54 & 76.10  &   70.70 &  69.69 &57.87      &80.14  & 70.04  & 77.35  &71.16   \\
     &CoT  & 77.58 &68.50 & 72.34 & 67.74 & 75.46  & 71.52   &  75.96 &65.50      &72.47  & 64.32  &73.87   & 67.99  \\
     \hline
       \multirow{7}{*}{\textbf{GT}} 
     &SF   & 71.52  &  66.28    & 78.40 & 69.87 &77.74& 70.98  & 69.34 & 62.97&  77.70& 67.83 & 76.31  &   68.19 \\
      &CW$_\text{soft}$   & 67.76  & 62.75     & 75.61 &  65.57 &75.12& 67.82  & 67.25 &60.45 & 73.87 & 61.65 &  73.52 &  64.10 \\
      &CW$_\text{hard}$   &  68.09 &   63.62   & 74.30 & 64.59  &74.47&67.74   & 68.99 &62.93 & 73.17 &  61.50&74.22& 66.01  \\
      &SBA$_\text{dir}$  & 73.16  &  67.96    &  79.21&  70.51 &79.87&73.06   & 72.82 &66.46 &  78.40& 68.11 & 78.40  &  69.80  \\
      &SBA$_\text{CoT}$ & \underline{78.07}  & \underline{68.32}    &71.85  & 67.83  & 76.92& 73.94   & \underline{78.40} &\underline{66.07} & 72.47 & 67.75 &  77.00 &   72.44 \\
      &SBA$_\text{exp}$ & 74.47  &  64.57    &80.03  & 70.13  &73.65& 68.98  &71.40&58.31 & 80.49 &  67.44 & 75.26 &  69.20 \\
      &SBA$_\text{ens}$  & 76.76  &  68.35    & 67.92 &  64.39 &66.61&  64.53 & 75.61 &65.19 & 67.94 &  63.13&68.29   &  64.79  \\
       \hline
       \multirow{7}{*}{\textbf{Hyb.}} 
     &SF  &  67.10 &    62.30  & 78.07 &  68.96 & 75.45&  68.62  & 64.46 & 58.72& 77.70 &65.65  &  74.91 &  66.31  \\
      &CW$_\text{soft}$&  70.05 &   65.00   & 77.41 &  68.35 & 77.25&70.02&  70.73&64.15 &  77.00&66.05  &  77.35 &  67.85   \\
      &CW$_\text{hard}$ &  70.38 &  65.49    &77.74  & 68.96  & 76.92& 69.72  &  70.38& 64.36& 77.35& 67.14 &  75.26 &  66.30   \\
      &SBA$_\text{direct}$ & 74.96  &  69.30    & 79.05 & 70.20  &\underline{80.03}&\underline{73.46}  &  74.91&68.06 & 78.75 & 68.06 &  \underline{78.75} &  \underline{70.76} \\
      &SBA$_\text{CoT}$& 75.29  &  65.78    &  73.00&  68.69 &75.29 & 71.90 &73.87&61.20 & 74.22 &  68.26& 75.96  &  71.14  \\
      &SBA$_\text{exp}$ & 76.10  & 64.53  & \underline{80.69} &  \underline{70.50} &72.83&  70.50 & 75.26 &60.88 & \underline{82.93}& \underline{70.56} &  74.91 &  71.47   \\
      &SBA$_\text{ens}$  & 76.76  &67.75 & 66.78 & 63.25  & 67.27  & 63.81  & 74.91 &63.39 & 64.11 & 58.01 & 66.55  &   60.07\\
    \hline
\end{tabularx}

  \caption{Performance of retrieval-augmented LLMs on the \textbf{ModC} and \textbf{HumC} splits of our CONFACT dataset. \textit{Baseline} (\textbf{Bsl.}) denotes models without incorporating source backgrounds.\textit{GT-MB} (\textbf{GT}) represents models that only consider incorporating source backgrounds with ground-truth human annotations. \textit{Hybrid-MB} (\textbf{Hyb.}) demonstrates models incorporated with both human-annotated source backgrounds as well as automatically generated media backgrounds. The best results (the highest summation of Acc. and F1) are underlined.}
  \label{tab:exp-results}
\end{table*}

\subsection{Main Experimental Results}
\label{sec:exp-result}
We conducted extensive experiments on the ModC and HumC splits of \textbf{CONFACT} (for implementation details, see Appendix~\ref{sec:app-exp-eval-setting}) to evaluate the performance of retrieval-augmented LLMs in fact-checking scenarios involving conflicting evidence. Our evaluation compares baseline RAG models that do not consider media source backgrounds (Baseline) against models that integrate source credibility data at different stages of the pipeline, using the strategies introduced in Section~\ref{sec:method-media-incorp} (i.e., Source Filtering (\textbf{SF}), Credibility Weighting (\textbf{CW$_{[\cdot]}$}), and Source Background Augmentation (\textbf{SBA$_{[\cdot]}$)}). The experiment results are presented in Table~\ref{tab:exp-results}. Below, we analyze key findings from our experiments by addressing three research questions.

\noindent\textbf{RQ 1}: \textit{How do vanilla retrieval-augmented LLMs perform when confronted with conflicting evidence from sources of varying credibility?}
  
As shown in the first block of Table~\ref{tab:exp-results}, vanilla RAG models exhibit difficulties when dealing with conflicting evidence. Their performance is notably limited, as reflected by lower F1 scores, suggesting challenges in correctly classifying the minority class (i.e., claims where the majority of retrieved evidence is misleading). 
This is primarily due to three key issues. First, hallucination — when presented with conflicting sources, LLMs sometimes generate factually incorrect responses that do not accurately reflect the retrieved evidence. Second, over-reliance on high-frequency responses — the Majority Vote setting biases the system toward the dominant source perspective, often amplifying misinformation if it is overrepresented in retrieval. Third, inability to distinguish misinformation from reliable sources — since vanilla RAG models do not assess source credibility, they treat all retrieved documents as equally valid, leading to incorrect fact-checking outputs. Notably, using GPT-4o in RAG methods (Appendix~\ref{sec:additiona_exp}) showed no clear advantage over open-source models, highlighting the problem's complexity.

Among the baseline answering strategies, Discern-and-Answer (\textbf{DisA.)} and Chain-of-Thought (\textbf{CoT}) prompting achieve better results than direct answer generation. This improvement suggests that prompting LLMs to explicitly reason about retrieved content helps mitigate the influence of unreliable sources. However, despite these improvements, the overall accuracy and F1 scores remain suboptimal, highlighting the need for more effective mechanisms to incorporate source credibility into the fact-checking process.

\noindent\textbf{RQ 2}: \textit{Does incorporating media source backgrounds improve fact-checking performance in RAG-based LLMs?}
 
Incorporating media background information into RAG models generally leads to improved  performance, although the degree of improvement varies across models.
Specifically, LLaMA-3.1 shows a 10\% absolute improvement in F1 score, while Mistral achieves a 5\% in accuracy improvement when media backgrounds are integrated on the ModC split. Similar improvements are observed on HumC, with the incorporation of source credibility information. These results indicate that providing source credibility cues helps LLMs resolve conflicting evidence more effectively.

However, not all models benefit equally from media backgrounds. Specifically, Qwen-2 exhibits the least improvement, which we attribute to its weaker long-context processing capabilities. The inclusion of source background information significantly increases input length. In models that do not handle extended sequences efficiently, this can dilute relevant context, increase token misalignment, and disrupt self-attention mechanisms, ultimately leading to suboptimal fact-checking performance. This finding suggests that as LLM architectures improve in handling long inputs, the benefits of integrating source-aware fact-checking will likely become more pronounced.

\noindent\textbf{RQ 3}: \textit{What is the most effective strategy for incorporating media source backgrounds into retrieval-augmented LLMs?}

Different strategies for integrating media backgrounds show distinct patterns of performance across RAG models. Our results indicate that the most effective approach is to incorporate media backgrounds at the answer generation stage, combined with a structured reasoning strategy such as \textbf{CoT} prompting or explicit instructions to discern unreliable source. 

In contrast, strategies that introduce media backgrounds in earlier stages-such as retrieval or ranking-are less effective. This is likely due to information loss when converting detailed textual source descriptions into a single credibility level or a credibility score. The credibility score predictor, despite being trained on expert-annotated data, does not always provide precise mappings between background descriptions and factual reliability, leading to potential misclassifications.

Furthermore, credibility-aware ranking strategies (\textbf{CW$_\text{soft}$} and \textbf{CW$_\text{hard}$}) sometimes degrade performance. This occurs because credibility and relevance are not always aligned—highly credible sources may not contain the most pertinent evidence for verifying a claim. Additionally, credibility-based filtering can risk removing crucial counter-evidence. Fact-checking often requires evaluating misleading claims in context, and aggressively filtering out sources deemed unreliable may leave models without the necessary contrastive information to identify misinformation. As a result, ranking methods that overly rely on credibility scores can paradoxically reduce fact-checking accuracy by limiting the model’s ability to reason over conflicting viewpoints.

Comparing GT-MB (which uses expert-verified MBFC credibility labels) and Hybrid-MB (which estimates credibility for missing sources using LLM-based retrieval), 
we do not observer obvious superiority of  Hybrid-MB.
This indicates that current source credibility estimation methods remain limited, which could add noise to source credibility aware RAG methods.
Manually curated credibility assessments are still more reliable than automated credibility prediction. Detailed error analysis is provided in Appendix~\ref{sec:app-error-analysis}.

\paragraph{Summary of Findings.}  
Our results demonstrate that retrieval-augmented LLMs struggle with conflicting evidence when source credibility is not explicitly considered. Integrating media backgrounds improves performance, but the effectiveness of this approach depends on how and where the information is introduced within the pipeline. The most effective strategy is incorporating background information at the answer generation stage, where structured reasoning techniques such as Chain-of-Thought prompting or  explicit instructions to discern unreliable source to resolve conflicting claims more effectively. In contrast, relying solely on credibility-aware filtering or ranking may inadvertently introduce biases or remove crucial context needed for fact-checking.

Our findings also reveal a fundamental trade-off between using expert-verified credibility data (GT-MB) and automated credibility estimation (Hybrid-MB). While expert annotations provide higher reliability, automated credibility inference allows for broader source coverage and scalability. Improving the accuracy of LLM-based credibility prediction remains a key open challenge for future research. These insights contribute to the broader field of AI-driven fact-checking by demonstrating both the potential and limitations of leveraging source credibility to enhance retrieval-augmented generation for misinformation detection.

\begin{table}[t]
  \centering
  \small
  \begin{tabularx}{\columnwidth}{X|X|X|XX|XX}
    \hline
   \multirow{2}{*}{\textbf{Top-$K$}} & \multirow{2}{*}{\textbf{Chk.}} & \multirow{2}{*}{\textbf{Meth.}} &
\multicolumn{2}{c|}{\textbf{LLaMA-3.1}} & \multicolumn{2}{c}{\textbf{Qwen-2}} \\ 
& && \textbf{Acc.} & \textbf{F1} & \textbf{Acc.} & \textbf{F1}  \\
   \hline
   \hline
    \multirow{4}{*}{Top-10} & \multirow{4}{*}{Para.}
     &SBA$_\text{dir}$ & 71.78  & 65.04 &  79.09&  69.76 \\
     & &SBA$_\text{CoT}$ & 76.66  & 66.02 & 57.14 &  67.34 \\
     & &SBA$_\text{exp}$ & 74,22  & 63.06&65.16  &  67.86 \\
     & &SBA$_\text{ens}$ & 74.56  &63.62  & 50.52 &  48.43 \\\hline
     \multirow{4}{*}{Top-5} & \multirow{4}{*}{Sent.} &SBA$_\text{dir}$ &  62.37 & 57.52 & 72.82 &  62.28 \\
    &  &SBA$_\text{CoT}$ & 67.94  & 58.84 & 61.32 &  62.28 \\
    & &SBA$_\text{exp}$ & 67.25  & 54.71 & 75.12 & 64.18 \\
    &  &SBA$_\text{ens}$ & 68.99  &59.76  & 54.01 & 53.90  \\
       \hline
\end{tabularx}

  \caption{Ablation results when using top-$10$ pieces of augmented context paragraph (para.) and top-$5$ sentence-level (sent.) chunking strategy.}
\label{tab:exp-ablations}
\end{table}

\subsection{Ablation Studies}
\label{sec:exp-ablations}
To further understand the impact of media source backgrounds on fact-checking with conflicting evidence, we conduct ablation studies focusing on GT-MB model on HumC, as they avoid noise from automated source estimation (Hybrid-MB) and HumC is more challenging as shown in Section~\ref{sec:exp-result}. Here, we consider the most powerful way (i.e., in the answer generation stage) to incorporate source credibility information.

\noindent\textbf{Effect of the Number of Augmented Paragraphs.}  
We assess whether increasing the number of retrieved paragraphs improves fact-checking performance by expanding the evidence set from 5 (Table~\ref{tab:exp-results}) to 10 documents (the first block in Table~\ref{tab:exp-ablations}). Surprisingly, this does not enhance accuracy as models may struggle with long inputs as well as be distracted from irrelevant information.

The main reasons are twofold: (1) Increasing the number of retrieved documents introduces lower-relevance evidence, which makes it harder for the model to discern factual correctness. (2) Longer input sequences overwhelm LLM attention mechanisms, leading to poorer factual reasoning. These findings suggest that retrieving fewer but more relevant documents is more effective than increasing retrieval breadth when dealing with conflicting claims.

\noindent\textbf{Impact of Chunking Strategies.}  
We compare paragraph-level (Table~\ref{tab:exp-results})  vs. sentence-level chunking (the second block in Table~\ref{tab:exp-ablations}) for retrieved evidence in the fact-checking pipeline. Paragraph-level chunking consistently outperforms sentence-level chunking, as fragmented sentences often lack sufficient context to resolve factual disputes. However, longer paragraph inputs increase computational overhead.  

A potential solution is de-contextualization methods, where sentences are supplemented with surrounding context before being processed by LLMs. Future work could explore such strategies to maintain high-context resolution while minimizing input length constraints.

\begin{table}[t]
\centering
\small

  \begin{tabular}{c|ccc }
    \hline
    \textbf{} &\textbf{w/o Background} & \textbf{GT} & \textbf{Hybrid} \\ \hline\hline
    \textbf{Acc.} & 49.45 &  \underline{\textbf{50.34}} &  48.47  \\        \hline
    \end{tabular}
    \caption{Human performance on CONFACT without background information, provided with GT background information, and hybrid background information.}
   \label{tab:exp-human}
\end{table}
\subsection{Human Evaluation}
\label{sec:exp-human-eval}
Beyond the quantitative analysis in Section~\ref{sec:exp-result}, we conduct a qualitative study to assess human fact-checking performance under conflicting evidence. We select 20 fact-checking claims from CONFACT and recruit four NLP researchers as human evaluators. Each human evaluator evaluates 10 claims across three settings, mirroring Section~\ref{sec:exp-result}: (1) without any source background, (2) with curated media backgrounds from MBFC (GT), and (3) with both MBFC-curated and automatically generated source backgrounds (Hybird). The accuracy of human evaluations is summarized in Table~\ref{tab:exp-human}.

Our findings indicate that humans often respond with "unsure" when faced with conflicting evidence, mirroring model performance: while GT media backgrounds boost accuracy, hybrid sources (including AI-generated backgrounds) tend to introduce noise and mislead evaluators. This suggests that unreliable or AI-generated context can impair judgment rather than enhance it.

These results have critical implications for real-world fact-checking organizations. Fact-checkers must adopt rigorous source verification methods to mitigate misinformation risks, and automated tools should prioritize high-fidelity data curation over broad retrieval to reduce misleading noise. Moreover, AI-generated evidence should be treated as assistive rather than authoritative, with human oversight ensuring effective verification of conflicting claims.

Overall, this human evaluation highlights the complexity of fact-checking amid conflicting evidence, reinforcing the need for high-quality evidence retrieval and robust verification mechanisms in both human and automated fact-checking systems.


\section{Conclusion}
This study presents a systematic evaluation of RAG models in fact-checking scenarios involving conflicting evidence—a critical yet underexplored challenge. To support this, we introduce the CONFACT dataset, which pairs fact-checking claims with contradictory information from sources of varying credibility. Our analysis indicates that existing RAG models struggle when faced with conflicting evidence, often ascribing undue reliability to less credible sources.

To address this issue, we integrate background information from the media sources into the RAG pipelines. Our findings reveal that incorporating source credibility signals during answer generation significantly enhances performance by reducing the models' susceptibility to misinformation. However, challenges remain, particularly in accurately assessing source credibility and mitigating biases in evidence retrieval.

These findings have practical implications for real-world fact-checking. Automated systems must go beyond naive retrieval and adopt rigorous source validation to avoid amplifying unreliable claims. Moreover, AI-assisted verification should complement human expertise, ensuring that models serve as tools to support rather than replace professional fact-checkers. Future work should focus on refining automated credibility assessments, improving evidence ranking, and reasoning under uncertainty.

Despite the progress demonstrated, we acknowledge several limitations of our current framework, including biases in source labeling and the absence of more advanced baseline systems. These are discussed in Appendix~\ref{sec:limitation}

Overall, our study confronts the complexities of conflicting evidence in fact-checking, and underscores the urgent need for trustworthy, AI-driven verification systems. Addressing these challenges is essential for strengthening resilience against misinformation and ensuring the reliability of AI-assisted fact-checking in journalistic, policy, and public discourse contexts.

\clearpage
\section*{Acknowledgement}
This research/project is supported by the National Research Foundation, Singapore under its National Large Language Models Funding Initiative (AISG Award No: AISG-NMLP-2024-004). Any opinions, findings and conclusions or recommendations expressed in this material are those of the author(s) and do not reflect the views of National Research Foundation, Singapore. This research/project is supported by the Ministry of Education, Singapore, under its SUTD-SMU Joint Grant Call, if applicable).

\bibliographystyle{named}
\bibliography{ijcai25}

\begin{thebibliography}{}

\bibitem[\protect\citeauthoryear{Amplayo \bgroup \em et al.\egroup }{2023}]{DBLP:conf/acl/AmplayoW00N23}
Reinald~Kim Amplayo, Kellie Webster, Michael Collins, Dipanjan Das, and Shashi Narayan.
\newblock Query refinement prompts for closed-book long-form {QA}.
\newblock In {\em Proceedings of the 61st Annual Meeting of the Association for Computational Linguistics (Volume 1: Long Papers), {ACL}}, pages 7997--8012, 2023.

\bibitem[\protect\citeauthoryear{Baly \bgroup \em et al.\egroup }{2018}]{DBLP:conf/emnlp/BalyKAGN18}
Ramy Baly, Georgi Karadzhov, Dimitar Alexandrov, James~R. Glass, and Preslav Nakov.
\newblock Predicting factuality of reporting and bias of news media sources.
\newblock In {\em Proceedings of the 2018 Conference on Empirical Methods in Natural Language Processing}, pages 3528--3539, 2018.

\bibitem[\protect\citeauthoryear{Baly \bgroup \em et al.\egroup }{2020}]{DBLP:conf/acl/BalyKAKDAGN20}
Ramy Baly, Georgi Karadzhov, Jisun An, Haewoon Kwak, Yoan Dinkov, Ahmed Ali, James~R. Glass, and Preslav Nakov.
\newblock What was written vs. who read it: News media profiling using text analysis and social media context.
\newblock In {\em Proceedings of the 58th Annual Meeting of the Association for Computational Linguistics, {ACL} 2020, Online, July 5-10, 2020}, pages 3364--3374, 2020.

\bibitem[\protect\citeauthoryear{Bashlovkina \bgroup \em et al.\egroup }{2023}]{DBLP:journals/corr/abs-2311-06697}
Vasilisa Bashlovkina, Zhaobin Kuang, Riley Matthews, Edward Clifford, Yennie Jun, William~W. Cohen, and Simon Baumgartner.
\newblock Trusted source alignment in large language models.
\newblock {\em CoRR}, abs/2311.06697, 2023.

\bibitem[\protect\citeauthoryear{Chen and Shu}{2024}]{DBLP:conf/iclr/ChenS24}
Canyu Chen and Kai Shu.
\newblock Can llm-generated misinformation be detected?
\newblock In {\em The Twelfth International Conference on Learning Representations, {ICLR}}, 2024.

\bibitem[\protect\citeauthoryear{Chen \bgroup \em et al.\egroup }{2021}]{DBLP:conf/acl/ChenGLL020}
Anthony Chen, Pallavi Gudipati, Shayne Longpre, Xiao Ling, and Sameer Singh.
\newblock Evaluating entity disambiguation and the role of popularity in retrieval-based {NLP}.
\newblock In {\em Proceedings of the 59th Annual Meeting of the Association for Computational Linguistics and the 11th International Joint Conference on Natural Language Processing, {ACL/IJCNLP}}, pages 4472--4485, 2021.

\bibitem[\protect\citeauthoryear{Chen \bgroup \em et al.\egroup }{2024}]{DBLP:conf/naacl/ChenKSDC24}
Jifan Chen, Grace Kim, Aniruddh Sriram, Greg Durrett, and Eunsol Choi.
\newblock Complex claim verification with evidence retrieved in the wild.
\newblock In {\em Proceedings of the 2024 Conference of the North American Chapter of the Association for Computational Linguistics: Human Language Technologies (Volume 1: Long Papers), {NAACL}}, pages 3569--3587, 2024.

\bibitem[\protect\citeauthoryear{Dubey \bgroup \em et al.\egroup }{2024}]{dubey2024llama}
Abhimanyu Dubey, Abhinav Jauhri, Abhinav Pandey, Abhishek Kadian, Ahmad Al-Dahle, Aiesha Letman, Akhil Mathur, Alan Schelten, Amy Yang, Angela Fan, et~al.
\newblock The llama 3 herd of models.
\newblock {\em arXiv preprint arXiv:2407.21783}, 2024.

\bibitem[\protect\citeauthoryear{Gao \bgroup \em et al.\egroup }{2023}]{DBLP:journals/corr/abs-2312-10997}
Yunfan Gao, Yun Xiong, Xinyu Gao, Kangxiang Jia, Jinliu Pan, Yuxi Bi, Yi~Dai, Jiawei Sun, Qianyu Guo, Meng Wang, and Haofen Wang.
\newblock Retrieval-augmented generation for large language models: {A} survey.
\newblock {\em CoRR}, abs/2312.10997, 2023.

\bibitem[\protect\citeauthoryear{Guo \bgroup \em et al.\egroup }{2022}]{DBLP:journals/tacl/GuoSV22}
Zhijiang Guo, Michael~Sejr Schlichtkrull, and Andreas Vlachos.
\newblock A survey on automated fact-checking.
\newblock {\em Trans. Assoc. Comput. Linguistics}, 10:178--206, 2022.

\bibitem[\protect\citeauthoryear{Guu \bgroup \em et al.\egroup }{2020}]{guu2020retrieval}
Kelvin Guu, Kenton Lee, Zora Tung, Panupong Pasupat, and Mingwei Chang.
\newblock Retrieval augmented language model pre-training.
\newblock In {\em International conference on machine learning}, pages 3929--3938. PMLR, 2020.

\bibitem[\protect\citeauthoryear{Hong \bgroup \em et al.\egroup }{2024}]{DBLP:conf/naacl/HongKKMW24}
Giwon Hong, Jeonghwan Kim, Junmo Kang, Sung{-}Hyon Myaeng, and Joyce~Jiyoung Whang.
\newblock Why so gullible? enhancing the robustness of retrieval-augmented models against counterfactual noise.
\newblock In {\em Findings of the Association for Computational Linguistics: {NAACL}}, pages 2474--2495, 2024.

\bibitem[\protect\citeauthoryear{Hounsel \bgroup \em et al.\egroup }{2020}]{DBLP:conf/uss/HounselHKBFM20}
Austin Hounsel, Jordan Holland, Ben Kaiser, Kevin Borgolte, Nick Feamster, and Jonathan~R. Mayer.
\newblock Identifying disinformation websites using infrastructure features.
\newblock In {\em 10th {USENIX} Workshop on Free and Open Communications on the Internet, {FOCI}}, 2020.

\bibitem[\protect\citeauthoryear{Kwon and others}{2023}]{vllm}
Woosuk Kwon et~al.
\newblock Efficient memory management for large language model serving with paged attention.
\newblock In {\em Proc. of the ACM SIGOPS 29th Symposium on Operating Systems Principles}, 2023.

\bibitem[\protect\citeauthoryear{Lee \bgroup \em et al.\egroup }{2024}]{DBLP:journals/corr/abs-2404-12447}
Yoonsang Lee, Xi~Ye, and Eunsol Choi.
\newblock Ambigdocs: Reasoning across documents on different entities under the same name.
\newblock {\em CoRR}, abs/2404.12447, 2024.

\bibitem[\protect\citeauthoryear{Lewis \bgroup \em et al.\egroup }{2020}]{DBLP:conf/nips/LewisPPPKGKLYR020}
Patrick S.~H. Lewis, Ethan Perez, Aleksandra Piktus, Fabio Petroni, Vladimir Karpukhin, Naman Goyal, Heinrich K{\"{u}}ttler, Mike Lewis, Wen{-}tau Yih, Tim Rockt{\"{a}}schel, Sebastian Riedel, and Douwe Kiela.
\newblock Retrieval-augmented generation for knowledge-intensive {NLP} tasks.
\newblock In {\em Advances in Neural Information Processing Systems 33: Annual Conference on Neural Information Processing Systems 2020, NeurIPS 2020, December 6-12, 2020, virtual}, 2020.

\bibitem[\protect\citeauthoryear{Li \bgroup \em et al.\egroup }{2023}]{DBLP:journals/corr/abs-2311-03731}
Dongfang Li, Zetian Sun, Xinshuo Hu, Zhenyu Liu, Ziyang Chen, Baotian Hu, Aiguo Wu, and Min Zhang.
\newblock A survey of large language models attribution.
\newblock {\em CoRR}, abs/2311.03731, 2023.

\bibitem[\protect\citeauthoryear{Min \bgroup \em et al.\egroup }{2020}]{DBLP:conf/emnlp/MinMHZ20}
Sewon Min, Julian Michael, Hannaneh Hajishirzi, and Luke Zettlemoyer.
\newblock Ambigqa: Answering ambiguous open-domain questions.
\newblock In {\em Proceedings of the 2020 Conference on Empirical Methods in Natural Language Processing, {EMNLP}}, pages 5783--5797, 2020.

\bibitem[\protect\citeauthoryear{Mistral.AI}{2023}]{misral7bv2}
Mistral.AI.
\newblock La plateforme, 2023.

\bibitem[\protect\citeauthoryear{Mukherjee and Weikum}{2015}]{DBLP:conf/cikm/MukherjeeW15}
Subhabrata Mukherjee and Gerhard Weikum.
\newblock Leveraging joint interactions for credibility analysis in news communities.
\newblock In {\em Proceedings of the 24th {ACM} International Conference on Information and Knowledge Management, {CIKM}}, pages 353--362, 2015.

\bibitem[\protect\citeauthoryear{Nakov \bgroup \em et al.\egroup }{2021}]{DBLP:conf/ijcai/NakovCHAEBPSM21}
Preslav Nakov, David P.~A. Corney, Maram Hasanain, Firoj Alam, Tamer Elsayed, Alberto Barr{\'{o}}n{-}Cede{\~{n}}o, Paolo Papotti, Shaden Shaar, and Giovanni Da~San Martino.
\newblock Automated fact-checking for assisting human fact-checkers.
\newblock In {\em Proceedings of the Thirtieth International Joint Conference on Artificial Intelligence, {IJCAI}}, pages 4551--4558, 2021.

\bibitem[\protect\citeauthoryear{Pan \bgroup \em et al.\egroup }{2023a}]{DBLP:conf/emnlp/PanLKN23}
Liangming Pan, Xinyuan Lu, Min{-}Yen Kan, and Preslav Nakov.
\newblock Qacheck: {A} demonstration system for question-guided multi-hop fact-checking.
\newblock In {\em Proceedings of the 2023 Conference on Empirical Methods in Natural Language Processing, {EMNLP}}, pages 264--273, 2023.

\bibitem[\protect\citeauthoryear{Pan \bgroup \em et al.\egroup }{2023b}]{DBLP:conf/acl/PanWLLWKN23}
Liangming Pan, Xiaobao Wu, Xinyuan Lu, Anh~Tuan Luu, William~Yang Wang, Min{-}Yen Kan, and Preslav Nakov.
\newblock Fact-checking complex claims with program-guided reasoning.
\newblock In {\em Proceedings of the 61st Annual Meeting of the Association for Computational Linguistics (Volume 1: Long Papers), {ACL}}, pages 6981--7004, 2023.

\bibitem[\protect\citeauthoryear{Popat \bgroup \em et al.\egroup }{2016}]{DBLP:conf/cikm/PopatMSW16}
Kashyap Popat, Subhabrata Mukherjee, Jannik Str{\"{o}}tgen, and Gerhard Weikum.
\newblock Credibility assessment of textual claims on the web.
\newblock In {\em Proceedings of the 25th {ACM} International Conference on Information and Knowledge Management, {CIKM}}, pages 2173--2178, 2016.

\bibitem[\protect\citeauthoryear{Popat \bgroup \em et al.\egroup }{2017}]{DBLP:conf/www/PopatMSW17}
Kashyap Popat, Subhabrata Mukherjee, Jannik Str{\"{o}}tgen, and Gerhard Weikum.
\newblock Where the truth lies: Explaining the credibility of emerging claims on the web and social media.
\newblock In {\em Proceedings of the 26th International Conference on World Wide Web Companion}, pages 1003--1012, 2017.

\bibitem[\protect\citeauthoryear{Ram \bgroup \em et al.\egroup }{2023}]{DBLP:journals/tacl/RamLDMSLS23}
Ori Ram, Yoav Levine, Itay Dalmedigos, Dor Muhlgay, Amnon Shashua, Kevin Leyton{-}Brown, and Yoav Shoham.
\newblock In-context retrieval-augmented language models.
\newblock {\em Trans. Assoc. Comput. Linguistics}, 11:1316--1331, 2023.

\bibitem[\protect\citeauthoryear{Robertson and Zaragoza}{2009}]{DBLP:journals/ftir/RobertsonZ09}
Stephen~E. Robertson and Hugo Zaragoza.
\newblock The probabilistic relevance framework: {BM25} and beyond.
\newblock {\em Found. Trends Inf. Retr.}, 3(4):333--389, 2009.

\bibitem[\protect\citeauthoryear{Schlichtkrull \bgroup \em et al.\egroup }{2023}]{DBLP:conf/nips/SchlichtkrullG023}
Michael Schlichtkrull, Zhijiang Guo, and Andreas Vlachos.
\newblock Averitec: {A} dataset for real-world claim verification with evidence from the web.
\newblock In {\em Advances in Neural Information Processing Systems 36: Annual Conference on Neural Information Processing Systems 2023, NeurIPS}, 2023.

\bibitem[\protect\citeauthoryear{Schlichtkrull}{2024}]{DBLP:journals/corr/abs-2409-00781}
Michael Schlichtkrull.
\newblock Generating media background checks for automated source critical reasoning.
\newblock {\em CoRR}, abs/2409.00781, 2024.

\bibitem[\protect\citeauthoryear{Thorne \bgroup \em et al.\egroup }{2018}]{thorne-etal-2018-fever}
James Thorne, Andreas Vlachos, Christos Christodoulopoulos, and Arpit Mittal.
\newblock {FEVER}: a large-scale dataset for fact extraction and {VER}ification.
\newblock In {\em Proceedings of the 2018 Conference of the North {A}merican Chapter of the Association for Computational Linguistics}, pages 809--819, 2018.

\bibitem[\protect\citeauthoryear{Wang \bgroup \em et al.\egroup }{2024a}]{DBLP:journals/corr/abs-2408-11871}
Lionel~Z. Wang, Yiming Ma, Renfei Gao, Beichen Guo, Zhuoran Li, Han Zhu, Wenqi Fan, Zexin Lu, and Ka~Chung Ng.
\newblock Megafake: {A} theory-driven dataset of fake news generated by large language models.
\newblock {\em CoRR}, abs/2408.11871, 2024.

\bibitem[\protect\citeauthoryear{Wang \bgroup \em et al.\egroup }{2024b}]{DBLP:conf/emnlp/WangWGZWXSWLQYL24}
Xiaohua Wang, Zhenghua Wang, Xuan Gao, Feiran Zhang, Yixin Wu, Zhibo Xu, Tianyuan Shi, Zhengyuan Wang, Shizheng Li, Qi~Qian, Ruicheng Yin, Changze Lv, Xiaoqing Zheng, and Xuanjing Huang.
\newblock Searching for best practices in retrieval-augmented generation.
\newblock In {\em Proceedings of the 2024 Conference on Empirical Methods in Natural Language Processing, {EMNLP}}, pages 17716--17736, 2024.

\bibitem[\protect\citeauthoryear{Wang}{2017}]{wang-2017-liar}
William~Yang Wang.
\newblock {\textquotedblleft}liar, liar pants on fire{\textquotedblright}: A new benchmark dataset for fake news detection.
\newblock In {\em Proceedings of the 55th Annual Meeting of the Association for Computational Linguistics}, pages 422--426, 2017.

\bibitem[\protect\citeauthoryear{Wei \bgroup \em et al.\egroup }{2022}]{DBLP:conf/nips/Wei0SBIXCLZ22}
Jason Wei, Xuezhi Wang, Dale Schuurmans, Maarten Bosma, Brian Ichter, Fei Xia, Ed~H. Chi, Quoc~V. Le, and Denny Zhou.
\newblock Chain-of-thought prompting elicits reasoning in large language models.
\newblock In {\em Advances in Neural Information Processing Systems 35: Annual Conference on Neural Information Processing Systems 2022, NeurIPS}, 2022.

\bibitem[\protect\citeauthoryear{Yang \bgroup \em et al.\egroup }{2024}]{yang2024qwen2}
An~Yang, Baosong Yang, Binyuan Hui, Bo~Zheng, Bowen Yu, Chang Zhou, Chengpeng Li, Chengyuan Li, Dayiheng Liu, Fei Huang, et~al.
\newblock Qwen2 technical report.
\newblock {\em arXiv preprint arXiv:2407.10671}, 2024.

\bibitem[\protect\citeauthoryear{Zhang and Gao}{2024}]{zhang-gao-2024-reinforcement}
Xuan Zhang and Wei Gao.
\newblock Reinforcement retrieval leveraging fine-grained feedback for fact checking news claims with black-box {LLM}.
\newblock In {\em Proceedings of the 2024 Joint International Conference on Computational Linguistics, Language Resources and Evaluation (LREC-COLING)}, pages 13861--13873, 2024.

\bibitem[\protect\citeauthoryear{Zhang \bgroup \em et al.\egroup }{2019}]{DBLP:conf/emnlp/ZhangMBRAKSJKBD19}
Yifan Zhang, Giovanni Da~San Martino, Alberto Barr{\'{o}}n{-}Cede{\~{n}}o, Salvatore Romeo, Jisun An, Haewoon Kwak, Todor Staykovski, Israa Jaradat, Georgi Karadzhov, Ramy Baly, Kareem Darwish, James~R. Glass, and Preslav Nakov.
\newblock Tanbih: Get to know what you are reading.
\newblock In {\em Proceedings of the 2019 Conference on Empirical Methods in Natural Language Processing and the 9th International Joint Conference on Natural Language Processing, {EMNLP-IJCNLP}}, pages 223--228, 2019.

\end{thebibliography}
\clearpage
\section*{APPENDIX}
\appendix
\section{Prompts for Annotation with GPT-4}
\label{sec:app-prompt-annotation}
\begin{tcolorbox}[colback=gray!10!white,colframe=black,title=A. System Prompt,fonttitle=\bfseries\small,fontupper=\small]
You are an expert in fact-checking. Analyze the claim, evidence, and claim date. Consider the timeline and disregard post-claim events. Determine if the evidence supports, rejects, or is inconclusive about the claim.
\end{tcolorbox}

\begin{tcolorbox}[colback=gray!10!white,colframe=black,title=B. URL Prompt,fonttitle=\bfseries\small,fontupper=\small]
Review the URL content to determine its support, rejection, or neutrality toward the claim. Consider the claim date. Respond only with: - Support - Reject - Not enough evidence. No additional text.
\begin{verbatim}
Claim: {claim}
Date of Claim: {claim_date}
URL: {evidence_url}
\end{verbatim}
\end{tcolorbox}
\begin{tcolorbox}[colback=gray!10!white,colframe=black,title=C. Text Prompt without Justification,fonttitle=\bfseries\small,fontupper=\small]
Review the text to determine its position on the claim considering the claim date. Respond only with: - Support - Reject - Not enough evidence. No additional text.
\begin{verbatim}
Claim: {claim}
Date when the claim was made: {claim_date}
Scraped Content: {evidence_content}
\end{verbatim}
\end{tcolorbox}
\begin{tcolorbox}[colback=gray!10!white,colframe=black,title=D. Text Prompt with Justification,fonttitle=\bfseries\small,fontupper=\small]
Evaluate the text against the claim date. Assess if it supports, rejects, or is inconclusive about the claim. Provide up to 500 words of reasoning and conclude with: - Support - Reject - Not enough evidence. Start your conclusion with 'Final answer: '.
\begin{verbatim}
Claim: {claim}
Date when the claim was made: {claim_date}
Scraped Content: {evidence_content}
\end{verbatim}
\end{tcolorbox}

In this section, we provide the specific prompts used for the first stage of conflicting evidence annotation with GPT-4 (Section~\ref{sec:dataset-conflicts}). We used three variants prompts to query GPT-4 in order to enhance the robustness of annotation. The three prompting strategies all leveraged the same system prompt as shown in Box A, and their individual prompting instructions are provided in Box B, C and D, respectively.

\section{Illustration of Data on \textsf{CONFACT}}
\label{sec:app-illu-data}
\begin{figure}[t] 
	\centering
	\includegraphics[width=0.95\linewidth]{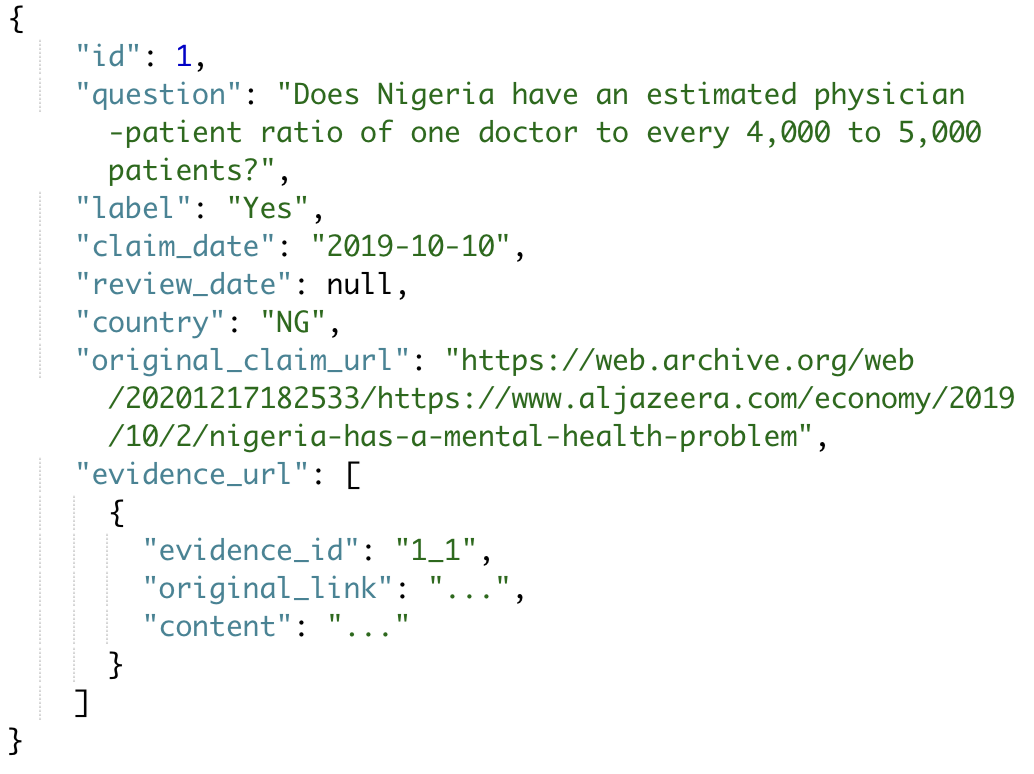} 
	\caption{A Data Sample from CONFACT}
	\label{fig:sample_data}
\end{figure}

Fig~\ref{fig:sample_data} presents a sample from the \textsf{CONFACT} dataset. Each instance in \textsf{CONFACT} comprises a question (converted from its original claim), a ground-truth answer, and a set of conflicting evidence.

\section{Implementation Details}
\label{sec:app-exp-eval-setting}
To handle long retrieved documents effectively, we apply a paragraph-based chunking strategy, where each document is split into passages of at most 256 words. This ensures that retrieved evidence remains contextually relevant while fitting within the token constraints of large language models. For retrieval and ranking, we use BM25~\cite{DBLP:journals/ftir/RobertsonZ09} to select the top-5 most relevant paragraphs for each claim. While we experimented with more advanced neural ranking approaches~\cite{DBLP:conf/nips/SchlichtkrullG023}, we did not observe any significant improvements in our setting.

\vspace{0.2em}
Inference is performed using the vLLM framework~\cite{vllm}, which optimizes key-value (KV) cache memory for efficient large-scale inference. We conducted all experiments with BFloat16 precision on a cluster of $2 \times$ NVIDIA A100-80 GPUs, employing greedy decoding for response generation. To evaluate RAG-based fact-checking performance, we experiment with three state-of-the-art LLMs: LLaMA-3.1-8B~\cite{dubey2024llama}, Qwen-2-8B~\cite{yang2024qwen2}, and Mistral-v0.3-7B~\cite{misral7bv2}. For methods that leveraging source credibility scores, we set the score threshold $\gamma$ to 0.3 and the balancing hyperparameter $\beta$ to 0.8, based on preliminary experiments optimizing both accuracy and macro-F1 performance.

\begin{table*}[ht!]
\centering
\small
 \begin{tabularx}{\textwidth}{X|XXXXXX}
 \hline
  & \makecell[c]{Academic/\\Research} 
  & \makecell[c]{Government} 
  & \makecell[c]{Mainstream \\News Media} 
  & \makecell[c]{Non-profit \\Organisation} 
  & \makecell[c]{Social Media} 
  & \makecell[c]{Others} \\ \hline \hline
 Very Unreliable & 2   & 0   & 0   & 0   & 3   & 3   \\ \hline
 Unreliable      & 10  & 1   & 27  & 5   & 25  & 34  \\ \hline
 Neutral         & 18  & 7   & 231 & 59  & 47  & 95  \\ \hline
 Reliable        & 65  & 74  & 334 & 110 & 15  & 112 \\ \hline
 Very Reliable   & 20  & 65  & 27  & 33  & 0   & 4   \\ \hline
 \end{tabularx}
 \caption{Reliability Breakdown for each Media Source Type}
 \label{tab:source-breakdown}
 \end{table*}

\FloatBarrier
\section{Limitations}
\label{sec:limitation}

We identify three main limitations of our study.

\vspace{0.2em}
\noindent\textbf{Bias in Source Credibility Annotations.}  
Our credibility annotations rely on the Media Bias/Fact Check (MBFC) dataset, which may carry inherent biases. As a result, any systematic bias present in MBFC is inherited by our framework and may influence model behavior during training and evaluation.

\vspace{0.2em}
\noindent\textbf{Context-Independent Credibility Assumptions.}  
Our approach treats media source credibility as static and context-independent. In reality, a source’s reliability may vary across topics, time periods, or issue-specific reporting. Future work could explore dynamic, context-aware credibility estimation methods to address this limitation.

\vspace{0.2em}
\noindent\textbf{Simplified Baselines.}  
While more sophisticated baselines that decompose claims into finer-grained components may improve factual verification, we deliberately adopt a simplified setting to isolate the effect of source credibility. Integrating decomposition-based approaches remains an important direction for future work.

\section{Error Analysis}
\label{sec:app-error-analysis}

We conduct an error analysis to examine the limitations of best performing RAG-based LLM (Llama-3.1) using GT-MB. We randomly sampled 50 cases where the model produced incorrect answers and categorized the errors. 

\vspace{0.2em}
\noindent\textbf{Errors Due to Conflicting or Irrelevant Retrieved Contexts.}  
These errors occur when the retrieved evidence either lacks relevance or presents conflicting claims. For example, in response to the question ``\textit{Has climate change increased hurricane frequency?}'', one retrieved source affirmed an increase, while another refuted it based on different datasets. Instead of reconciling the conflicting claims, the model incorrectly aligned with the source that contained more surface-level keyword matches. This suggests that simple retrieval-based approaches struggle with conflicting evidence, reinforcing the need for advanced ranking methods that assess both relevance and credibility before generating an answer. Future work could explore iterative ranking and selection strategies where the model answers a question only when a sufficient set of corroborative evidence is identified.

\vspace{0.2em}
\noindent\textbf{Errors from Inaccurate Media Background Estimation.}  
This error type is prevalent in the Hybrid-MB setting, where automatically generated media backgrounds misclassify source reliability, leading the model to prioritize misleading information. For instance, in answering ``\textit{Has the deficit come down under the Conservatives?},'' the ground truth answer is \textit{Yes}, supported by data from Full Fact, a highly credible source. However, retrieved evidence from Tax Research UK, a critical government watchdog, suggested \textit{No}. Due to an erroneous background classification labeling Tax Research UK as highly reliable, the model was misled and answered incorrectly. These findings highlight the risks of relying on generated media backgrounds without rigorous validation, emphasizing the necessity of robust source credibility estimation techniques.

\noindent\textbf{LLM Bias in Resolving Conflicting Evidence.}  
Some errors stem from the model’s tendency to favor the majority viewpoint among retrieved contexts, even when the opposing evidence is more credible. For example, in cases where three low-credibility sources supported one claim while a single authoritative source contradicted them, the model frequently defaulted to the majority position. This bias suggests that LLMs may lack the ability to critically weigh conflicting evidence, underscoring the need for improved reasoning mechanisms that incorporate source reliability assessments. Mitigating such biases requires integrating structured fact-checking methodologies that encourage LLMs to assess the credibility of competing claims rather than defaulting to frequency-based heuristics.

Overall, these error categories reveal fundamental challenges in fact-checking with conflicting evidence. Addressing them requires advancements in retrieval ranking, media background estimation, and bias mitigation to ensure automated fact-checking systems align with real-world journalistic and verification practices.

\section{Distribution of Source Credibility}
\label{sec:app-dis-source-cred}

Table~\ref{tab:source-breakdown}
shows the credibility annotations across different source types. We observed over-estimation of credibility for mainstream news media by non-expert annotators. They have annotated about $\mathbf{95\%}$ of mainstream news media as trustworthy probably due to their popularity. However, journalism have identified $\mathbf{30\%}$ of them as incredible.

\section{Detailed Implementations for RAG Methods}
\label{sec:app-detail-rag}

\subsection{Credibility Related Information Extraction from Background}
\label{sec:app-cred-ext-from-background}
\begin{tcolorbox}[colback=gray!10!white,colframe=black,title=E. Prompt to Classify Credibility,fonttitle=\bfseries\small,fontupper=\small]
You are InfoHuntGPT, a world-class AI assistant used by journalists to predict the credibility of media sources. Your task is to read the example media sources (with their corresponding credibility descriptions) and then assess the target media using only the details provided. Do not include any additional information, and follow these rules strictly:

\begin{enumerate}
    \item Read all the information about each example media source and note its final credibility rating.
    \item Examine the target media’s description and any relevant Wikipedia or article information, if provided.
    \item Determine the target media’s credibility using one of these labels: \texttt{high}, \texttt{medium}, or \texttt{low}.
    \begin{itemize}
        \item \textbf{Low}: The source demonstrates questionable reliability, has a track record of publishing misinformation, failed fact checks, or lacks transparency.
        \item \textbf{Medium}: The source is generally reliable but may display occasional bias or minor factual issues, with no consistent pattern of publishing false information.
        \item \textbf{High}: The source consistently provides reliable, factual information, uses proper sourcing, and shows little to no history of failed fact checks.
    \end{itemize}
    \item Pay special attention to any indication of failed fact checks or repeated misinformation—this may reduce credibility.
    \item If there is virtually no information or the details appear highly suspicious, you may conclude \texttt{low} credibility.
    \item Output only the final credibility assessment for the target media (one word: \texttt{high}, \texttt{medium}, or \texttt{low}).
    \item Do not provide disclaimers or references to your reasoning. Do not include any additional commentary. Provide the credibility label in one line.
\end{enumerate}

\textbf{Example 1}\\
Media Description: [Media Background Details]\\
Wikipedia: [Wikipedia Summary]\\
Credibility: [High/Medium/Low]\\
...\\
\textbf{Example N}\\
...\\
 \\
\textbf{Target Media Description}: [Media Background Details]\\
\textbf{Target Media Wikipedia}: [Wikipedia Summary]\\
\textbf{Target Media Credibility}:
\end{tcolorbox}

Predicted source background descriptions are detailed and comprehensive, whereas contain irrelevant information about the source credibility (e.g., the founder of a source). Considering the issue, we further distill the credibility related content from the description using the prompt illustrated in Box E. 
Specifically, we classify the credibility of a source into three levels: \textit{low}, \textit{medium} and \textit{high}, following ~\cite{DBLP:conf/emnlp/BalyKAGN18}.
To ensure the accuracy of our automatic source credibility prediction framework, we further conducted a quantitative evaluation on~\cite{DBLP:conf/emnlp/BalyKAGN18}. Our automatic source credibility prediction framework achieved 70.04\%
in a zero-shot manner.

\subsection{Source Background Incorporation}

\label{sec:app-src-back-incorp}
\begin{tcolorbox}[colback=gray!10!white,colframe=black,title=F. System Prompt,fonttitle=\bfseries\small,fontupper=\small]
You are given a question and several pieces of evidence. Your task is to analyze the evidence and provide a concise answer to the question. \\
For each piece of evidence, the background of its source media is provided. When evaluating the evidence, it is crucial to take into account the credibility of the source media, as this can significantly influence the reliability of the evidence. Additionally, consider any potential biases that may be inherent in the source media, especially if they are explicitly mentioned. This will help ensure a more nuanced and thorough evaluation of the evidence, factoring in both the content and the context in which it is presented.

\end{tcolorbox}
\begin{tcolorbox}
[colback=gray!10!white,colframe=black,title= G. SBA$_\text{dir}$,fonttitle=\bfseries\small,fontupper=\small]
\textbf{Evidence 1:} [Evidence text] \\
\textbf{Source Media Description:} [Description of the source media] \\
\ldots \\
\textbf{Evidence N:} [Evidence text] \\
\textbf{Source Media Description:} [Description of the source media]\\
\textbf{Question:} [Question to be answered based on the evidence provided]
\end{tcolorbox}

\begin{tcolorbox}
[colback=gray!10!white,colframe=black,title=H. SBA$_\text{CoT}$,fonttitle=\bfseries\small,fontupper=\small]
\textbf{Question:} [Question to be answered based on the evidence provided]\\
\textbf{Evidence 1:} [Evidence text] \\
\textbf{Source Media Description:} [Description of the source media] \\
\ldots \\
\textbf{Evidence N:} [Evidence text] \\
\textbf{Source Media Description:} [Description of the source media]\\

Given the above evidence, first explain your reasoning for any contradictions or conflicting information. After your reasoning, provide your final answer to the question. 
Start your answer with 'Final Answer' and clearly separate it from the rest of your analysis.
Your final answer should be either 'yes' or 'no'. 
Include only one final answer, and avoid adding any additional explanation after it.\\
\textbf{Output format:} \\
\texttt{Analysis: [Your reasoning here]} \\
\texttt{\#*\# Final Answer: [yes/no]}
\end{tcolorbox}

\begin{tcolorbox}
[colback=gray!10!white,colframe=black,title=I. SBA$_\text{exp}$,fonttitle=\bfseries\small,fontupper=\small]
\textbf{Question:} [Question to be answered based on the evidence provided]\\

Some evidence below may have been perturbed with wrong information. Find the perturbed passages and ignore them when eliciting the correct answer.

\textbf{Evidence 1:} [Evidence text] \\
\textbf{Source Media Description:} [Description of the source media] \\
\ldots \\
\textbf{Evidence N:} [Evidence text] \\
\textbf{Source Media Description:} [Description of the source media]\\

First, thoroughly analyze all the provided evidence before making your final decision. Identify the perturbed sentences and carefully consider their implications in your analysis. Once you have completed your review, provide your final answer to the question based on the evidence you analyzed. Start your answer with 'Final Answer:' and ensure it is clearly separated from your evidence analysis. \textbf{Your final answer should be either 'yes' or 'no’. Make sure to include only one final answer, and do not include any additional text after it.}
\end{tcolorbox}

When incorporating source background in the generation stage of RAG methods, for the SBA$_\text{dir}$, SBA$_\text{CoT}$ and SBA$_\text{exp}$ settings, the source-aware  paragraphs (i.e., concatenation of the paragraph and the background description of its source) serve as the augmented context for answer generation, while using different instructions. 
In SBA$_\text{CoT}$, the model is required to provide a rationale besides the answer; in SBA$_\text{exp}$ the model is explicitly instructed to ignore augmented texts from incredible sources. Detailed prompts are shown in Box G and H.

\begin{figure}[t] 
	\centering
	\includegraphics[width=0.95\linewidth]{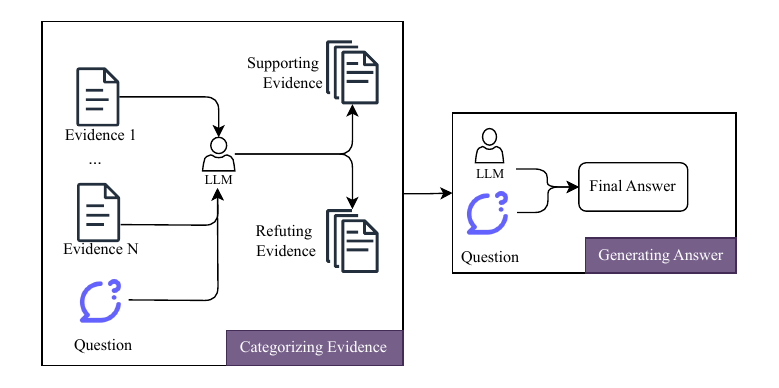} 
	\caption{Workflow of the Ensemble Method.}
	\label{fig:agent_based}
\end{figure}

For the SBA$_\text{ens}$ setting, a two stage prompting mechanism is exploited. As illustrated in Fig \ref{fig:agent_based}, we first employ a LLM to categorize each piece of evidence as either supporting or refuting (Box J). Next, we prompt the LLM to generate the final answer using the categorized evidence (Box K).

\begin{tcolorbox}
[colback=gray!10!white,colframe=black,title=J. SBA$_\text{ens}$: Categorizing Evidence,fonttitle=\bfseries\small,fontupper=\small]
For each piece of evidence, the background of its source media is provided. When evaluating the evidence, it is crucial to take into account the credibility of the source media, as this can significantly influence the reliability of the evidence. Additionally, consider any potential biases that may be inherent in the source media, especially if they are explicitly mentioned. This will help ensure a more nuanced and thorough evaluation of the evidence, factoring in both the content and the context in which it is presented.\\
\textbf{Question:}[Question to Answer]\\
\textbf{Supporting Evidence:}\\
- Sentence: [Sentence/Paragraph]\\
- Credibility Analysis: [Description of the source media]\\
…\\
\textbf{Refuting Evidence:}\\
- Sentence: [Sentence/Paragraph]\\
- Credibility Analysis: [Description of the source media]\\
…\\
Given the above support and oppose evidence, first explain your reasoning for any contradictions or conflicting information. Your analysis should be no more than 500 words. Please ignore the difference in the amount of supporting and opposing evidence and choose more detailed and truthful sentences of evidence. Once you have completed your analysis, provide your final answer to the question based on the evidence you analyzed. Start your answer with 'Final Answer:' and ensure it is clearly separated from your evidence analysis. Your final answer should be either 'yes' or 'no'. Make sure to include only one final answer, and do not include any additional text after it.
\end{tcolorbox}

\begin{tcolorbox}
[colback=gray!10!white,colframe=black,title=K. SBA$_\text{ens}$: Generating Final Answer,fonttitle=\bfseries\small,fontupper=\small]
\textbf{Instructions}

\begin{enumerate}
  \item \textbf{Comprehend the Question:}
  \begin{itemize}
    \item Carefully read the question to understand what is being asserted.
    \item Identify the key components and assertions within the question.
  \end{itemize}

  \item \textbf{Analyze the Sentence:}
  \begin{itemize}
    \item Examine the sentence to see how it relates to the question.
    \item Determine if the sentence provides evidence, an example, or a counterpoint to the question.
    \item Look for keywords or phrases that directly support or refute the question.
  \end{itemize}

  \item \textbf{Evaluate the Media Background:}
  \begin{itemize}
    \item Review the media background information to understand the broader context.
    \item Consider the credibility of the sources mentioned and any potential biases.
    \item Identify any historical information or prior events that relate to the question.
  \end{itemize}

  \item \textbf{Integrate Information:}
  \begin{itemize}
    \item Combine insights from the sentence and media background.
    \item Assess whether the sentence, in the context of the media background, provides sufficient support for the question.
    \item Consider if there are contradictions or alignments between the sentence and the media background.
  \end{itemize}

  \item \textbf{Logical Reasoning:}
  \begin{itemize}
    \item Use critical thinking to evaluate the connections.
    \item Ask yourself if the evidence logically leads to the conclusion stated in the question.
    \item Consider alternative interpretations or whether additional information is needed.
  \end{itemize}

  \item \textbf{Conclude:}
  \begin{itemize}
    \item Evaluate the reliability of the media background and determine whether the sentence supports the question.
    \item Ensure that your conclusion is based solely on the information provided.
  \end{itemize}

  \item \textbf{Answer:}
  \begin{itemize}
    \item Optionally, provide a justification based on the above steps, explaining your reasoning. Keep your justification under 300 words.
    \item Provide a clear and concise \texttt{Yes} or \texttt{No} answer to the question.
  \end{itemize}
\end{enumerate}
\noindent
\textbf{Question:} [Question to Answer] \\
\textbf{Evidence:} [Evidence] \\
\textbf{Media Background Analysis:} [Source Media Description of the Evidence Provided]

\vspace{1em}
Based on the provided sentence and the media background, begin by thoroughly analyzing the evidence, giving special attention to the credibility and potential biases of the media source. After your analysis, provide your final answer to the question. Start your answer with 'Final Answer:'.

\end{tcolorbox}

\section{Annotation Instructions}
\label{sec:app-annotation-guide}
\begin{figure}[ht!]
    \centering
    \includegraphics[width=0.95\linewidth]{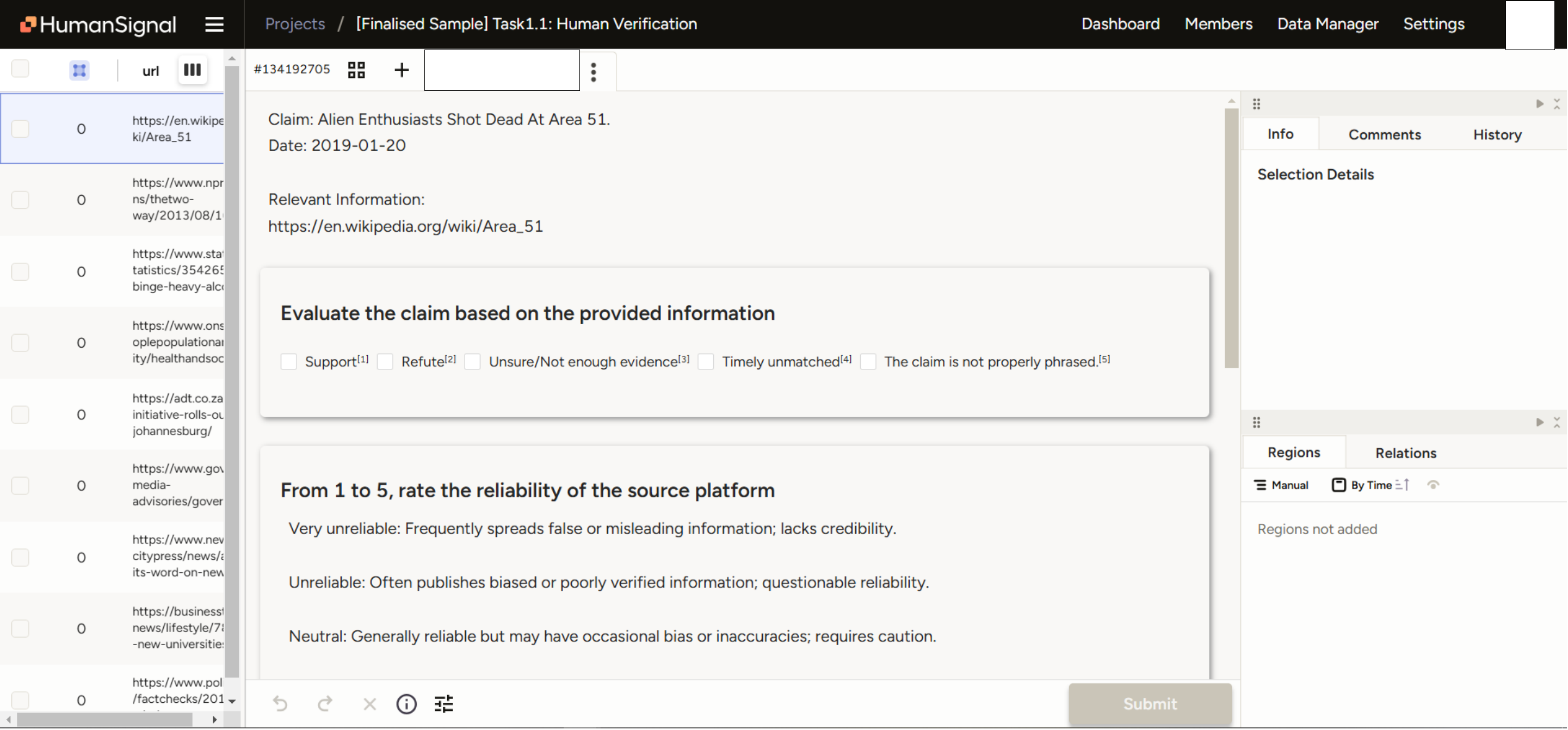}
    \caption{Screenshot of Annotation Platform}
    \label{fig:annotation-platform}
\end{figure}

Participants were presented with a claim and a corresponding URL linking to an online article or post. They were instructed to visit these URLs, review the content, and assess whether it supported or contradicted the claim. Following this evaluation, they were required to rate the source’s credibility and classify its type. Figure \ref{fig:annotation-platform} illustrates a screenshot of the user labeling process.

\section{Media Background Prediction}
\label{sec:app-media-background-pred}

\subsection{Demonstrations for Media Background Prediction. }

\begin{figure}[t] 
	\centering
	\includegraphics[width=0.95\linewidth]{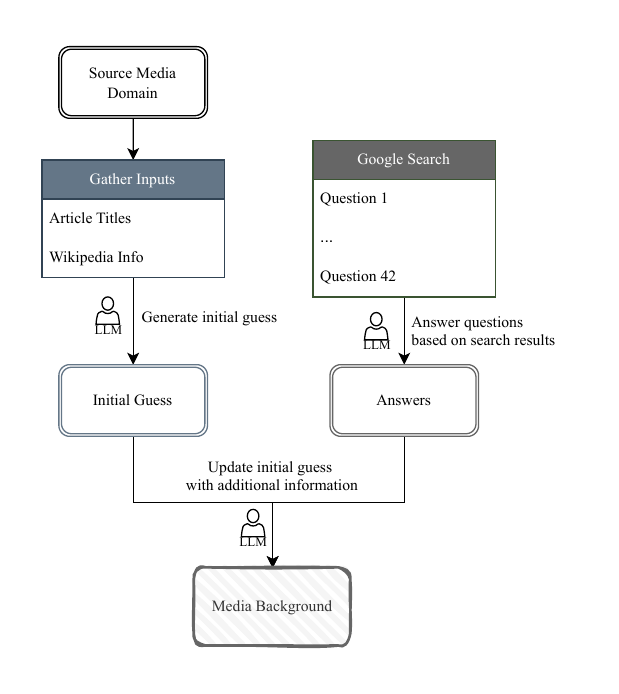} 
	\caption{Illustration of the process to generate the media background by LLMs.}
	\label{fig:mediaBG}
\end{figure}
As shown in Fig \ref{fig:mediaBG}, to build a media background for an unknown media source, we begin by gathering foundational information: specifically, we collect 10 article titles from the source and compile its Wikipedia summary to form an initial guess. Next, following the method in \cite{DBLP:journals/corr/abs-2409-00781}, we conduct a structured Google search using 42 targeted queries about the media, collecting the resulting answers to supplement our initial information. Finally, we incorporate these newly acquired insights into the initial hypothesis to produce a comprehensive, final media background description. An example of predicted media background is shown in Box L.

\begin{tcolorbox}[colback=gray!10!white,colframe=black,title=L. Generated Description for the Source Media,fonttitle=\bfseries\small,fontupper=\small]
\textbf{History:} Oxford University Press OUP is one of the oldest and largest university presses in the world, with a history dating back to 1586. \texttt{academic.oup.com} is the online platform for OUP that hosts a vast collection of scholarly journals, books, and articles. 

\vspace{0.5em}
\textbf{Funded by/Ownership:} 
OUP is a department of the University of Oxford and is self-funded through the sales of its publications. 
\vspace{0.5em}

\textbf{Analysis/Bias:} 
The platform, \texttt{academic.oup.com}, publishes peer-reviewed journals and books from a wide range of disciplines, including humanities, social sciences, and natural sciences. The editorial process for these publications is rigorous and follows the highest standards of academic integrity. While the platform may have a slight bias towards the perspectives of its authors and editors, it strives to provide balanced and informative content. 

\vspace{0.5em}
\textbf{Failed Fact Checks:} 
Oxford University Press has no record of any failed fact-checks. They are considered a reputable source of academic information. They have over 300 journals and a range of books that cover most disciplines.

\end{tcolorbox}

\subsection{Credibility Score Prediction}
\label{sec:app-credible-score}
We utilize the MBFC dataset to train a model for predicting media credibility scores. The dataset consists of media descriptions paired with credibility scores. 

The BigBird-RoBERTa model serves as the backbone of our architecture, combined with a regression head to output credibility scores. After training on the MBFC dataset, the model is evaluated and used to predict credibility scores for new media descriptions.

\section{Additional Experiment Results}
\label{sec:additiona_exp}

\begin{table}[t]
  \centering
  \small
  \begin{tabularx}{\columnwidth}{X|X|XX}
    \hline
   \multirow{2}{*}{\textbf{Set.}} & \multirow{2}{*}{\textbf{Meth.}} &
\multicolumn{2}{c}{\textbf{4o-mini}}\\ 
& & \textbf{Acc.} & \textbf{F1}\\
   \hline
   \hline
   \textbf{Bal.}& CoT & 78.23 & 70.42 \\
   \textbf{GT}& CoT & 79.05 & 71.33 \\
   \hline
\end{tabularx}
\caption{Additional Experiment Results using the 4o-mini Model}
   \label{tab:extra-exp}
\end{table}

We conducted additional experiments using the GPT-4o-mini model under the top 5 paragraphs setting. The results are presented in Table~\ref{tab:extra-exp}.
The results showed the conflicting evidence in fact-checking is also challenging to strong closed-source LLMs, indicating the need for source-critical fact-checking methods.

\end{document}